\documentclass[runningheads]{llncs}
\usepackage{graphicx}
\usepackage{tikz}
\usepackage{comment}
\usepackage{amsmath,amssymb} 
\usepackage{color}
\usepackage{graphicx}
\usepackage{float} 
\usepackage{amsmath}
\usepackage{amssymb}
\usepackage{booktabs}
\usepackage{mathrsfs}
\usepackage{multirow}
\usepackage{makecell}
\usepackage[width=122mm,left=12mm,paperwidth=146mm,height=193mm,top=12mm,paperheight=217mm]{geometry}

\begin{document}
\pagestyle{headings}
\mainmatter

\title{PanoViT: Vision Transformer for Room Layout Estimation from a Single Panoramic Image} 

\titlerunning{PanoViT: Vision Transformer for Room Layout Estimation}
\author{Weichao Shen \and
Yuan Dong \and
Zonghao Chen \and Zhengyi Zhao \and Yang Gao 
\and Zhu Liu}
\institute{Alibaba Group}

\maketitle

\begin{abstract}
In this paper, we propose PanoViT, a panorama vision transformer to estimate the room layout from a single panoramic image.  Compared to CNN models, our PanoViT is more proficient in learning global information from the panoramic image for the estimation of complex room layouts. Considering the difference between a perspective image and an equirectangular image, we design a novel recurrent position embedding and a patch sampling method for the processing of panoramic images. In addition to extracting global information, PanoViT also includes a frequency-domain edge enhancement module and a 3D loss to extract local geometric features in a panoramic image.
Experimental results on several datasets demonstrate that our method outperforms state-of-the-art solutions in room layout prediction accuracy.
\keywords{Room layout estimation, Panorama vision transformer, Panoramic image}
\end{abstract}

\section{Introduction}

Estimating 3D room layout from images plays an important role in many applications, such as virtual/augmented reality and robot vision. Compared with a perspective image with a narrow field of view (FOV), a 360-degree panoramic image contains the content of the entire room and provides more information for room layout estimation. With the popularity of 360 cameras, estimating the 3D room layout from a single panoramic image has become an active research topic \cite{zou20193d,pintore2020state}.

Recent works utilize the convolutional neural network (CNN)\cite{zou2018layoutnet,sun2019horizonnet,pintore2020atlantanet,sun2021hohonet,wang2021led2} to estimate the 3D room layout.
Although many methods have achieved good reconstruction accuracy, CNN-based methods still struggled in estimating complex room layouts.  One reason is that convolutional filters are more suitable at extracting local features, but estimating complex room layouts also requires global information in a panoramic image to deal with the occluded regions. In recent years, transformer \cite{vaswani2017attention} 
is very popular in many natural language processing (NLP) tasks \cite{devlin2018bert,radford2019language,brown2020language} and also shows strong performance on many computer vision tasks\cite{dosovitskiy2020image,carion2020end,liu2021swin}. Compared with CNNs, transformer is skilled at extracting the global relationship in an image, which is very useful for the estimation of complex room layouts. Therefore, in this article, we focus on 
introducing a vision transformer in estimating the room layout from a single panoramic image. 

There are several issues that need to be addressed when designing a vision transformer for panoramic images. First of all, as shown in many existing works, visual transformers need to be pre-trained on large data sets in order to work well. However, the spherical geometric information in panoramic images makes them different from perspective images, so that visual transformers pre-trained on perspective image datasets cannot be used to obtain satisfactory room layout estimation results from panoramic images. At the same time, the data in
panoramic image datasets is insufficient for adequate pre-training of a vision transformer.
The second problem is that the position attribute of the patch on a panoramic image is different from that on a perspective image, so that the position embedding methods for perspective images are not suitable for panoramic images.

In this article, we propose PanoViT,  a panorama vision transformer for estimating the room layout from a single RGB panoramic image. Since the visual transformer pre-trained on perspective images does not perform well on panoramic images, we introduce the CNN features in PanoViT as a bridge between the panoramic image and the visual transformer. Specifically, our model contains a CNN backbone to extract multi-scale features, then PanoViT takes the original panoramic image and the multi-scale features as the inputs. The CNN pre-trained on perspective image datasets performs well in processing panoramic images \cite{sun2021hohonet,wang2021led2}, so that the PanoViT can better process panoramic images by learning the global relationship among the panoramic image patches and the multi-scale CNN features. Considering the different position attributes between the patches sampled from panoramic images and the patches sampled from perspective images, we design a 
novel patch sampling method and a recurrent position embedding in PanoViT for processing panoramic images. The recurrent position embedding can ignore the absolute starting position of the patches in the horizontal direction, thereby emphasizing the translation invariance of a panoramic image in the horizontal axis. 

In addition to extracting global information, we also emphasize the ability of PanoViT to extract local features in panoramic images. Therefore, we design a frequency-domain edge enhancement module and 3D loss for PanoViT by exploiting the geometric information of panoramic images.  On the one hand, the edge information in a panoramic image is important for room layout estimation\cite{zou2018layoutnet}. 
We analyze the Fourier transform of panoramic images and propose an efficient edge enhancement method in the frequency domain. On the other hand, we design a 3D loss based on the geometric information of a panoramic image to measure the reconstruction error in 3D space, thereby alleviating the error imbalance problem caused by the calculation of the loss function in the image coordinates.

We characterize our contributions as follows:
\begin{itemize}
    \item We propose the PanoViT model to introduce vision transformers into the room layout estimation from a single RGB panoramic image. The experimental results on several datasets demonstrate that our method outperforms state-of-the-art solutions in room layout prediction accuracy.
    \item We designed a new patch sampling method and a recurrent position embedding for the vision transformer to process panoramic images. 
    \item We further design an edge enhancement method in the frequency domain and propose a 3D loss to use the geometric information in a panoramic image to improve the overall performance.
\end{itemize}
\section{Related works}
Estimating room layout from images has been a hot research topic for many years.  In this section, we only discuss the literature that focuses on the room layout estimation from a single panorama. We also analyze the works about vision transformers that are related to our method.

\paragraph{Room layout estimation from panoramic images} Compared with a perspective image, a panoramic image can capture a wide indoor environment that is useful for room layout estimation. Therefore, many room layout estimation methods rely on panoramic images. Zhang \emph{et al.} \cite{zhang2014panocontext} proposed to estimate the 3D bounding box of a room from an input panorama and constructed an annotated PanoContect dataset. Some extended works focused on improving the model with more geometry metric \cite{xu2017pano2cad,yang2016efficient} and semantic features \cite{yang2018automatic}. Recently, many methods 
utilized deep neural networks to improve room layout estimation. Zou \emph{et al.} \cite{zou2018layoutnet} designed the LayoutNet to directly predict the corner and boundary map from a panorama. Yang \emph{et al.} \cite{yang2019dula} proposed the DuLa-Net that can leverage different clues in both the equirectangular panorama-view image and the perspective ceiling-view image to predict the room layout. Unlike the dense prediction methods for layout estimation, Sun \emph{et al.} \cite{sun2019horizonnet} proposed the HorizonNet to leverage the property of aligned panoramic image to predict the positions of floor-wall, ceiling-wall boundaries, and wall-wall boundaries. Sun \emph{et al.} \cite{sun2021hohonet} further extended the HorizonNet with a latent horizontal features and proposed the HohoNet with much better speed and accuracy. Pintore \emph{et al.} \cite{pintore2020atlantanet} proposed the AtlantaNet to reconstruct rooms that are not following Manhattan world constraints. Wang \emph{et al.} \cite{wang2021led2} formulated the task of 360 layout estimation as a problem of predicting the depth on the horizon line of a panorama. Unlike all the existing methods based on CNN neural network, we propose a room layout estimation method based on vision transformers. Compared with CNN, our PanoViT can leverage the transformer to exploit the global vision dependencies in a panoramic image, which is helpful to the inference of the occluded room layout. 

\paragraph{Vision transformer} Transformer models have demonstrated excellent performance on a broad range of language tasks. The breakthroughs from transformer networks in NLP motivate a lot of works to adapt the transformer into vision tasks. Carion \emph{et al.} \cite{carion2020end} proposed a detect transformer(DETR) that firstly introduced the transformer into the visual detection task. Based on DETR, Zhu \emph{et al.} \cite{zhu2020deformable} proposed the deformable DETR to combine the best of the sparse spatial sampling of deformable convolution and the relation modeling capability of transformers. Recently, transformers were further introduced in different vision tasks, such as image classification \cite{dosovitskiy2020image}, semantic segmentation \cite{SETR}, and image generation \cite{esser2021taming}. All the existing vision transformer methods are designed to deal with the perspective images. While the 360 degree panoramic image processing becomes more popular, we attempt to introduce transformers in this area. 

\begin{figure*}[t]
\centering
    \includegraphics[width=0.95\textwidth]{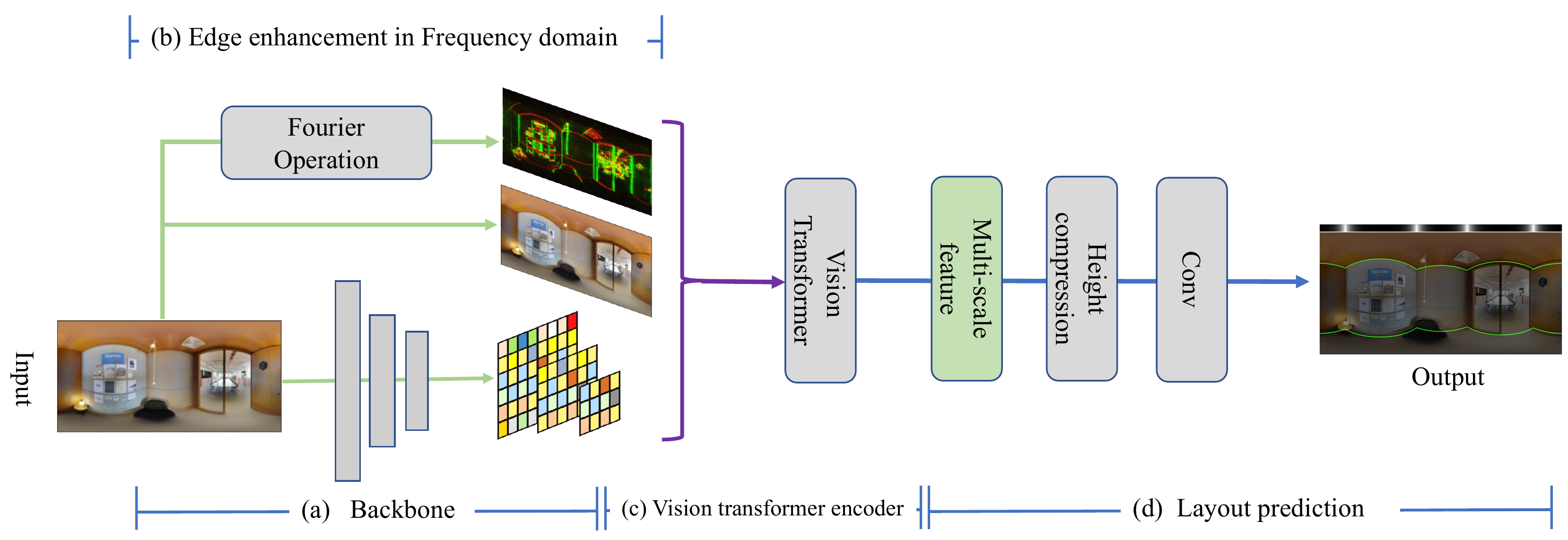}
    \caption{The overview of PanoViT. Our model can be divided into four modules, including (a) a backbone, (b) an edge enhancement module, (c) a vision transformer encoder, and (d) a layout prediction module.}
    \label{fig:framework}
\end{figure*}

\section{Approach}
\subsection{Overview}
The goal of the PanoViT model is to predict the room layout based on 360-degree panoramic images.
According to HorizonNet \cite{sun2019horizonnet}, we represent the room layout as a one-dimensional representation. Specifically, our model takes a 360-degree panoramic image as input and outputs a $C\times 1 \times W$ tensor, where $C = 3$ is the number of channels, and $W$ represents the width of the panoramic image. The first two channels represent the ceiling-wall and floor-wall boundaries, and the last one represents the probability of existence of the wall-wall boundary. Based on this representation, the framework of PanoViT's network can be divided into four modules, including a backbone, a panoramic visual transformer encoder, an edge enhancement module and a layout prediction module. A panoramic image is sent to the backbone to extract multi-scale feature maps, and sent to the edge enhancement module to obtain an edge enhancement map. The panoramic vision transformer encoder takes the original image, edge enhancement maps and multi-scale feature maps as input and outputs a feature vector for the layout prediction module to estimate the one-dimensional room layout representation. The pipeline of our network is shown in Fig.  \ref{fig:framework}. We will introduce each module in detail in the next section.

\subsection{Backbone}
The backbone is a convolutional network used to extract multi-scale features from panoramic images. In this article, we use ResNet34 as the backbone, and combine intermediate feature maps from different ResNet blocks as multi-scale features to capture low-level and high-level visual patterns. At the end of each block in ResNet34, we add a strip pooling layer \cite{hou2020strip} to make multi-scale features more robust.

\begin{figure*}[t]
\centering
    \includegraphics[width=0.9\textwidth]{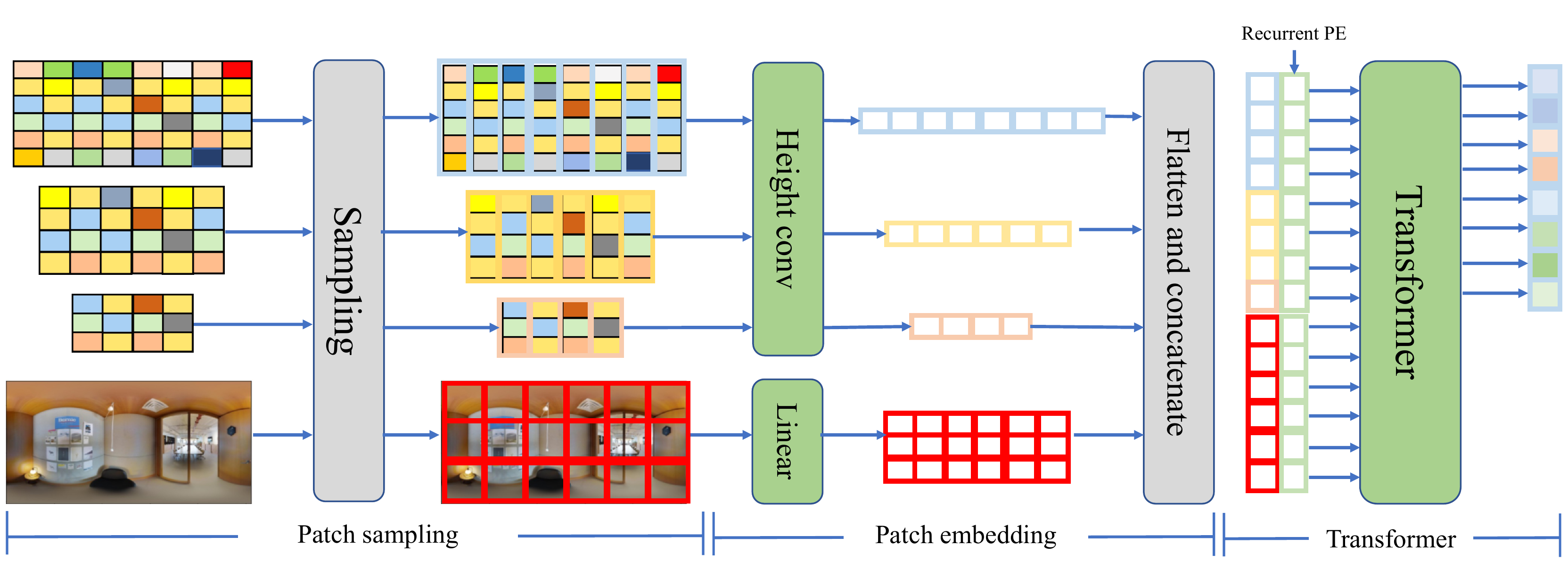}
    \caption{The overview of our panorama vision transformer. Our model contains a patching sampling module, a patch embedding module, and a vision transformer backbone.}
    \label{fig:transformer}
\end{figure*}

\subsection{Vision transformer encoder}
The vision transformer encoder takes the original image and multi-scale feature maps as input and outputs a room layout feature vector.
Fig. \ref{fig:transformer} shows the overview of the vision transformer encoder. The encoder first samples patches from the original image and the multi-scale feature maps. Then different patches are projected into different patch embeddings through different embedding operations. All the patch embeddings are attached with a recurrent position embedding to predict the room layout feature vector with a vision transformer backbone. 

\paragraph{Patch sampling}
We designed two different sampling methods for the multi-scale feature map and the original image. For multi-scale feature maps, we sample vertical patches along its horizontal axis. Specifically, given a feature map with a dimension of $C \times W \times H$, we sample it as $W$ patches of size $C \times 1 \times H$, where $H$ is the height of the original feature map, $W$ is the width of the original feature map, and $C$ is the number of channels. Feature maps of other scales are sampled in the same way. For panoramas, we sample square patches uniformly across the entire image. Given a panoramic image of size $C_{p} \times W_{p} \times H_{p}$, we reshape it into $N$ patches of $C_{p} \times P \times P$ , where $P $ is the size of the patch, and $N = W_{p}H_{p}/P^{2}$ is the number of patches.
\begin{figure*}[t]
\centering
    \includegraphics[width=0.90\textwidth]{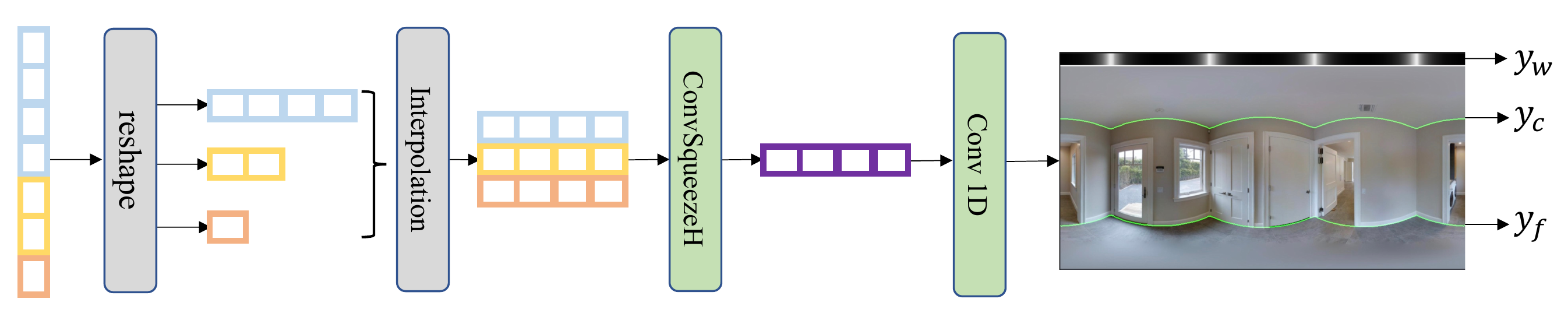}
    \caption{The overview of the layout prediction module.}
    \label{fig:layout_prediction}
\end{figure*}

\paragraph{Patch embedding}
We project different types of patches into patch embeddings through different embedding operations.
Inspired by the one-dimensional features used in \cite{sun2021hohonet}, we utilize a height compression module to calculate the patch embeddings of patches sampled from multi-scale feature maps. The height compression module is a convolutional layer containing $C_{emb}$ filters with kernel size $1 \times H$ to cover the height of feature maps. For a patch with the dimension of $C \times 1 \times H$, the 
height compression module project it into a one-dimensional patch embedding with the size of $C_{emb} \times 1 \times 1$. 
Notice that $C_{emb}$ does not change with the number of patch channels $C$, so that the one-dimensional patch embeddings have the same feature dimension $C_{emb}$ for patches sampled from feature maps with different scales. For patches sampled from the panoramic image, we use a simple multilayer perceptron to calculate their patch embeddings. All patch embeddings are concatenated together as the input of the vision transformer encoder.

\begin{figure*}[t]
\centering
    \includegraphics[width=0.7\textwidth]{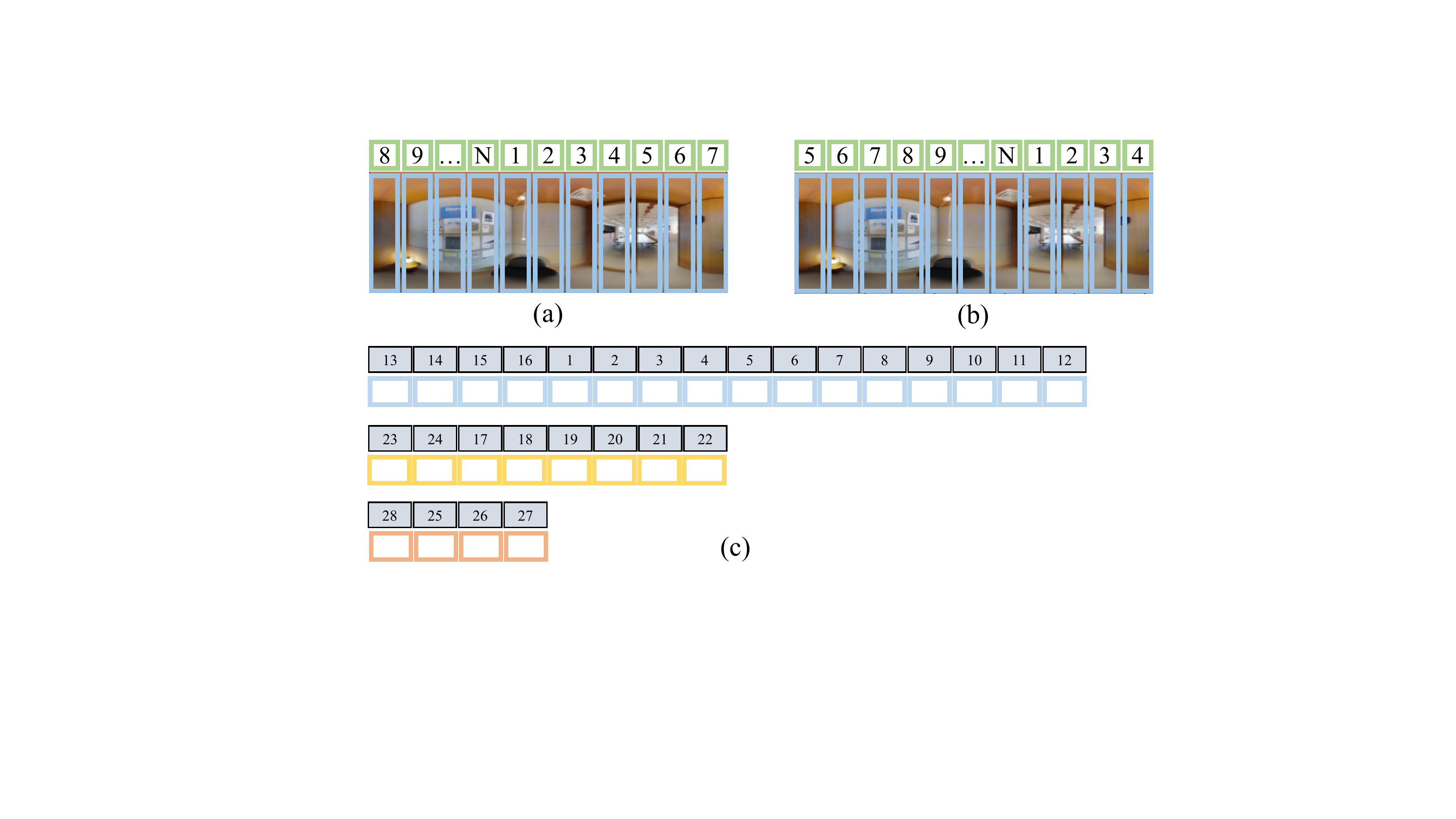}
    \caption{The presentation of the recurrent position embedding.
    In each image, the bottom row shows patches and the top row shows the corresponding position embedding. Figure (a) and (b) present two position embedding with different start positions. Figure (c) shows the recurrent position embedding for multi-scale features. The colored rectangles in different rows in figure (c) represent patches sampled from feature maps of different scales.}
    \label{fig:recurrent_PE}
\end{figure*}

\paragraph{Recurrent position embedding}
Existing position embeddings are all designed for perspective images. In this section, we design a recurrent position embedding for panoramic images. A major difference between a perspective image and a panoramic image is that the panoramic image is rolling stable, which means that the panoramic image rolling with any pixels along the horizon axis results in the same 3D layout compared to the original image. This property indicates that there is no need to emphasize the absolute starting position of patches along the horizontal axis. For instance, as shown in Fig. \ref{fig:recurrent_PE}, the position embedding shown in (a) should have the same result compared with (b). To emphasize this attribute, the recurrent position embedding is equipped with a starting position randomly sampled along the horizontal axis. In particular, our recurrent position embedding is calculated as
\begin{eqnarray} \label{Eq:1}
 \begin{aligned}
    RPE(pos,2i) &= sin\Big( \frac{pos+randinit}{10000^{2i/d_{model}}}\Big),\\
    RPE(pos,2i+1) &= cos\Big( \frac{pos+randinit}{10000^{2i/d_{model}}}\Big),
 \end{aligned}
\end{eqnarray} 
where $d_{model}$ is the dimension of the position embedding, $pos$ is the absolute position that starts from the left and ends at the right, and $randinit$ is a random initial position. 

For the multi-scale feature outputted from the ResNet, we compute the recurrent position embedding as following,
\begin{eqnarray} \label{Eq:2}
 \begin{aligned}
    RPE^{k}(pos,2i) &= sin\Big( \frac{pos+randinit+\sum_{k} s_{k}}{10000^{2i/d_{model}}}\Big),\\
    RPE^{k}(pos,2i+1) &= cos\Big( \frac{pos+randinit+\sum_{k} s_{k}}{10000^{2i/d_{model}}}\Big),\\
 \end{aligned}
\end{eqnarray} 
where the $RPE^{k}$ is the k-\emph{th} position embedding of the patches sampled from the feature map of the k-th ResNet block, and $s_{k}$ is the number of the patches sampled from the k-th feature map (according to our sampling method, $s_{k}$ equals to the width of the feature map from the k-th ResNet block).

\paragraph{Vision transformer backbone}
We use the ViT\cite{dosovitskiy2020image} pre-trained on the ImageNet as our backbone of the vision transformer. Compared to ViT, two changes in the transformer encoder are that we delete the extend patch in ViT designed for object classification, and we combine the feature vector of all the patches sampled from the multi-scale feature maps as the outputted room layout feature vector, whose length equals to $\sum^{N}_{k=1}(k)$ where $N$ is the number of the scales.

\subsection{Layout prediction}
The structure of the layout prediction module is illustrated in Fig. \ref{fig:layout_prediction}. We first reshape the room layout feature vector back to multi-scale features following the scale of the corresponding feature maps. The features with different scales are resized to a same length with the interpolation operation, and are combined together in the vertical dimension. Then we utilize the ConvSqueezeH layer introduced in \cite{sun2021hohonet} to compress the multi-scale features in the vertical dimension. Finally, we apply three Conv1D layers of kernel size 3, 3, and 1 respectively with BN, ReLU to output the probability $y_{w}$ of existence of the wall-wall boundary, the ceiling-wall $y_{c}$ and the floor-wall boundaries $y_{f}$.

\begin{figure*}[t]
\centering
    \includegraphics[width=0.95\textwidth]{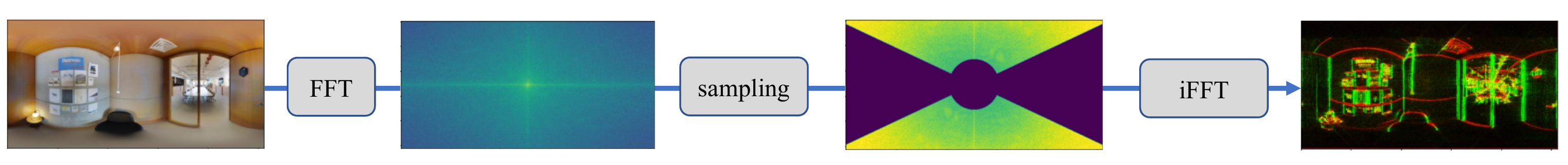}
    \caption{The overview of edge enhancement in frequency domain. We first compute the frequency transformation of a panoramic image with FFT. Then we design a filter according to the direction of the basis function to sample the frequency transformation and reconstruct the edge enhancement map with the iFFT.}
    \label{fig:fourier-pipeline}
\end{figure*}

\subsection{Edge enhancement}
The edge information is useful for the wall line detection, as shown in \cite{zou2018layoutnet}. In this paper, we extract the edge information in the frequency domain. Compared to the LSD used in \cite{zou2018layoutnet}, our method is more efficient and can achieve better accuracy. 

The edge enhancement module can be represented by 
\begin{eqnarray} \label{Eq:3}
 \begin{aligned}
    E = \mathscr{F}^{-1}(M\mathscr{F}(I))
 \end{aligned}
\end{eqnarray} 
where $I$ is a panorama, $E$ is the edge enhanced image, $M$ is a binary mask, $\mathscr{F}$ is the Fast Fourier transformation, and $\mathscr{F}^{-1 }$ is the inverse fast Fourier transformation.
Fig. \ref{fig:fourier-pipeline} shows the pipeline of the edge enhancement in the frequency domain. We first compute the frequency transformation of a panoramic image $P$ with the fast Fourier transformation $\mathscr{F}$. Then we sample a part of the frequency components with a mask $M$. Finally, we reconstruct the edge enhance map $E$ with the inverse Fourier transformation $\mathscr{F}^{-1}$.

\begin{figure*}[t]
\centering
    \includegraphics[width=0.8\textwidth]{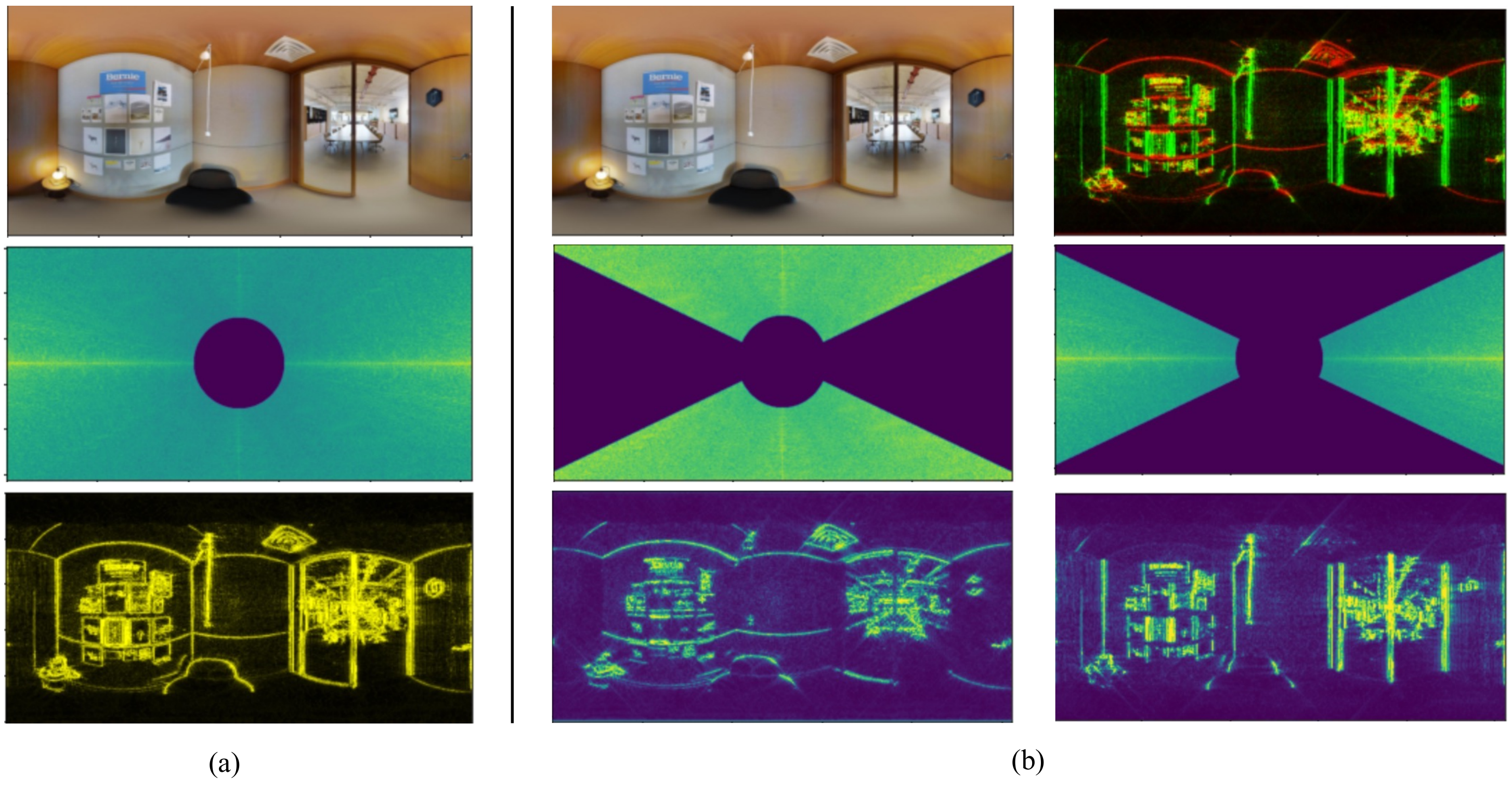}
    \caption{The presentation of different kinds of sampling strategies.
    From the top to the bottom, figure (a) shows the panoramic image, the high-pass filter, and the reconstructed edge enhancement map. From the left to the right and the top to the bottom, figure (b) presents the panoramic image, the reconstructed edge enhancement map, the horizon mask $M_{h}$, the vertical mask $M_{v}$, the corresponding reconstructed edge enhancement map of $M_{h}$, and the corresponding reconstructed edge enhancement map of $M_{v}$. }
    \label{fig:fourier_vis}
\end{figure*}
The key to edge enhancement is sampling different parts of the frequency transformation with the mask $M$. As is known, the high-frequency part contains the edge information while the low-frequency contains the mean content of the image. Therefore, a straightforward design for the mask $M$ is a high pass filter, and its reconstruction result is shown in Fig. \ref{fig:fourier_vis} (a). As shown in \cite{zou2018layoutnet}, separately extracting the edge along different axes (\emph{i.e.} x,y,z) helps to improve the accuracy. To extract the edge along different axes in the frequency domain, we propose to design the mask following the direction of the trigonometric function in the frequency transformation. 
The direction of the trigonometric function in the frequency transformation at $(u_{i},v_{i})$ can be calculated by 
\begin{equation}
\label{Eq:4}
    D_{i} = arctan(u_{i}-u_{0}/v_{i}-v_{0})
\end{equation}
where $(u_{0},v_{0})$ is the center of the frequency transformation. Trigonometric functions in different directions highlight the edges of the panorama in the same direction. Therefore, in order to obtain edges in different directions, we design two masks for edge enhancement in the horizontal and vertical directions. The calculation formulas are respectively  
\begin{eqnarray} \label{Eq:5}
 \begin{aligned}
M_{v}(u_{i},v_{i}) = \left\{
\begin{aligned}
&1,    &    &|D_{i}|<\alpha \& \sqrt{u_{i}^{2}+v_{i}^{2}} > \theta\\
&0,    &   &else\\
\end{aligned}
\right.
 \end{aligned}
\end{eqnarray} 
and
\begin{eqnarray} \label{Eq:6}
 \begin{aligned}
M_{h}(u_{i},v_{i}) = \left\{
\begin{aligned}
&1,    &    &|D_{i}|>=\beta \& \sqrt{u_{i}^{2}+v_{i}^{2}} > \theta\\
&0,    &    &else\\
\end{aligned}
\right.
 \end{aligned}
\end{eqnarray} 
where $\sqrt{u_{i}^{2}+v_{i}^{2}}> \theta$ works as a high pass filter, and both $\alpha$ and $\beta$ are hyperparameters. $M_{v}$ is the mask to extract the horizon edge enhanced image $E_{v}$, so we only maintain the trigonometric function whose direction angle is less than $\alpha$. The detailed information is shown in Fig. \ref{fig:fourier_vis} (b).
Finally, we concatenate two edge enhancement maps $E_{v}$ and $E_{h}$ with the original panoramic image $P$ in the channel dimension.

\begin{figure*}[t]
\centering
    \includegraphics[width=0.8\textwidth]{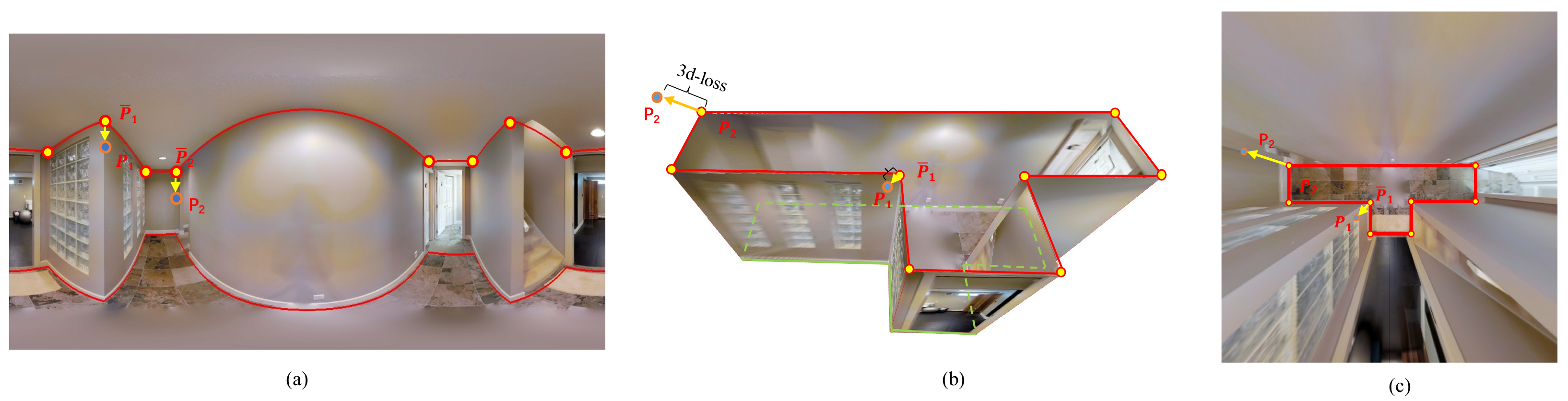}
    \caption{An illustration of the 3D loss. Figure (a) shows the ground truth boundary in red lines, a couple of the points $\overline{P}_{1}$ and $\overline{P}_{2}$ on the ground truth boundary, and the corresponding predicted points ${P}_{1}$ and ${P}_{2}$. In figure (a), points $P_{1}$ and $P_{2}$ have the same L1 error in the equirectangular image space, \emph{i.e.}, $\Delta(\overline{P}_{1} - {P}_{1})=\Delta(\overline{P}_{2} - {P}_{2})$. However, as illustrated figure (b) and figure (c), in the 3D space, $\Delta(\overline{P}_{1} - {P}_{1}) \neq \Delta(\overline{P}_{2} - {P}_{2})$. Compared to L1 loss in the equirectangular image space, our 3D loss can deal with the unbalance errors caused by the panoramic geometry.}
    \label{fig:TDloss}
\end{figure*}
\subsection{Loss functions}
The loss function consists of $\mathscr{L}_{cor}$ and $\mathscr{L}_{bon}$, where $\mathscr{L}_{cor}$ measures the error between the predicted position of the corners and the ground truths. We use the binary cross-entropy to define $\mathscr{L}_{cor}$. 

$\mathscr{L}_{bon}$ can be directly defined upon the errors of our two predicted ceiling and floor boundaries (i.e., $y_{f}$ and $y_{c}$) with respect to the ground truth. Previous works \cite{sun2019horizonnet,sun2021hohonet} always define this error in the equirectangular image space, which does not consider the geometry information in the panorama image. In this paper, we design a panoramic geometry-aware loss in 3D space (i.e. 3D loss) inspired by RenderLoss in L2DNet \cite{wang2021led2}. In particular, we represent a pixel q on the equirectangular image with its longitude $\delta$ and latitude $\gamma$, i.e., q = ($\delta$, $\gamma$).  Then we convert the pixel q to a $3$D point $P(p_x,p_y,p_z)$ in the $3$D space by
\begin{equation}
 \begin{aligned}
    &P(p_x,p_y,p_z) = \mathcal{F}(\delta,\gamma),\\
    &p_x = s \cdot cos(\gamma)\cdot sin(\delta),\\
    &p_y = s \cdot sin(\gamma),\\
    &p_z = s \cdot cos(\delta)\cdot cos(\gamma),\\
 \end{aligned}   
\end{equation}
where $\mathcal{F}$ denotes the convert function, $\delta$ $\in$ [-$\pi$,$\pi$], $\gamma$ $\in$ [-$\pi$/2,$\pi$/2], and $s$ denotes the scale which can be calculated by the predefined camera height. The 3D loss can be expressed as the $L_{1}$ error between the 3D point $P$ sampled from the boundary (\emph{i.e.}, $y_f$ and $y_c$) and the corresponding 3D point $\hat{P}$ sampled from the ground truth. An illustration of 3D loss is shown in Fig. \ref{fig:TDloss}.

\section{Experiments}
In this section, we test the performance of our method on different datasets and compare it with the state-of-the-art baselines. We also provide several ablation studies to analyze the effectiveness of different modules in our model.

\subsection{Dataset}
We conduct the experiments on two different datasets, including the PanoContext and the Matterport3D (Mp3D) dataset.
PanoContext dataset contains around 500 panoramas along with ground truth layout annotations.
The Matterport3D dataset contains $2,295$ panoramas of the complex cases with different numbers of layout corners. We adopt the official train/val/test split in \cite{zou20193d}. 

\subsection{Training Setup}
Our model is trained end-to-end with the Adam \cite{kingma2014adam} optimizer, where the learning rate is $0.0001$. The hyperparameters $\alpha$, $\beta$ and $\theta$ are set to $20$, $25$ and $100$. The batch size is set to $8$ and the epoch is set to $300$. During the training stage, we apply the same data augmentations following \cite{zou2018layoutnet}, including flip, rotation, gamma transformation, and stretch.
In addition, we design a new data augmentation method, adding $50$ random masks to the panoramic image, each of which is a $50\times 50$ rectangle with a value of $0$.

\begin{table*}
\centering
\caption{The results on the MP3D dataset. Our method achieves best results on both 2D IoU and 3D IoU.}
\label{tab:1}
\begin{tabular}{|c|c|c|ccccc|}
\hline
                          &                          &                          & \multicolumn{5}{c|}{\# of corners}                                                                                                                                                                                                                                                                  \\ \cline{4-8} 
\multirow{-2}{*}{Dataset} & \multirow{-2}{*}{Method} & \multirow{-2}{*}{Metric} & \multicolumn{1}{c|}{overall}                                 & \multicolumn{1}{c|}{4}                                       & \multicolumn{1}{c|}{6}                                       & \multicolumn{1}{c|}{8}                                       & 10+                                     \\ \hline
                          & AtlantaNet\cite{pintore2020atlantanet}               &                          & \multicolumn{1}{c|}{82.09\%}                                 & \multicolumn{1}{c|}{84.42\%}                                 & \multicolumn{1}{c|}{83.85\%}                                 & \multicolumn{1}{c|}{76.97\%}                                 & 73.18\%                                 \\ \cline{2-2} \cline{4-8} 
                          & HorizonNet \cite{sun2019horizonnet}             &                          & \multicolumn{1}{c|}{81.71\%}                                 & \multicolumn{1}{c|}{84.67\%}                                 & \multicolumn{1}{c|}{84.82\%}                                 & \multicolumn{1}{c|}{73.91\%}                                 & 70.58\%                                 \\ \cline{2-2} \cline{4-8} 
                          & HoHoNet \cite{sun2021hohonet}                  &                          & \multicolumn{1}{c|}{82.52\%}                                 & \multicolumn{1}{c|}{85.57\%}                                 & \multicolumn{1}{c|}{84.89\%}                                 & \multicolumn{1}{c|}{75.72\%}                                 & 70.88\%                                 \\ \cline{2-2} \cline{4-8} 
                          & LED2  \cite{wang2021led2}                     &                          & \multicolumn{1}{c|}{83.91\%}                                 & \multicolumn{1}{c|}{86.91\%}                                 & \multicolumn{1}{c|}{85.53\%}                                 & \multicolumn{1}{c|}{{\color[HTML]{FE0000} \textbf{78.72\%}}} & 71.79\%                                 \\ \cline{2-2} \cline{4-8} 
                          & Ours                 & \multirow{-5}{*}{2D IoU}  & \multicolumn{1}{c|}{{\color[HTML]{FE0000} \textbf{84.25\%}}} & \multicolumn{1}{c|}{{\color[HTML]{FE0000} \textbf{86.92\%}}} & \multicolumn{1}{c|}{{\color[HTML]{FE0000} \textbf{86.71\%}}} & \multicolumn{1}{c|}{{\color[HTML]{000000} 78.07\%}}          & {\color[HTML]{FE0000} \textbf{73.88\%}} \\ \cline{2-8} 
                          & AtlantaNet\cite{pintore2020atlantanet}              &                          & \multicolumn{1}{c|}{80.02\%}                                 & \multicolumn{1}{c|}{82.09\%}                                 & \multicolumn{1}{c|}{82.08\%}                                 & \multicolumn{1}{c|}{75.19\%}                                 & 71.61\%                                 \\ \cline{2-2} \cline{4-8} 
                          & HorizonNet \cite{sun2019horizonnet}             &                          & \multicolumn{1}{c|}{79.11\%}                                 & \multicolumn{1}{c|}{81.88\%}                                 & \multicolumn{1}{c|}{82.26\%}                                 & \multicolumn{1}{c|}{71.78\%}                                 & 68.32\%                                 \\ \cline{2-2} \cline{4-8} 
                          & HoHoNet \cite{sun2021hohonet}                  &                          & \multicolumn{1}{c|}{80.08\%}                                 & \multicolumn{1}{c|}{82.90\%}                                 & \multicolumn{1}{c|}{82.28\%}                                 & \multicolumn{1}{c|}{73.85\%}                                 & 69.17\%                                 \\ \cline{2-2} \cline{4-8} 
                          & LED2  \cite{wang2021led2}                      &                          & \multicolumn{1}{c|}{81.52\%}                                 & \multicolumn{1}{c|}{84.22\%}                                 & \multicolumn{1}{c|}{83.22\%}                                 & \multicolumn{1}{c|}{{\color[HTML]{FE0000} \textbf{76.89\%}}} & 70.09\%                                 \\ \cline{2-2} \cline{4-8} 
\multirow{-10}{*}{Mp3D}   & Ours                 & \multirow{-5}{*}{3D IoU} & \multicolumn{1}{c|}{{\color[HTML]{FE0000} \textbf{82.04\%}}} & \multicolumn{1}{c|}{{\color[HTML]{FE0000} \textbf{84.40\%}}} & \multicolumn{1}{c|}{{\color[HTML]{FE0000} \textbf{84.67\%}}} & \multicolumn{1}{c|}{{\color[HTML]{000000} 76.34\%}}          & {\color[HTML]{FE0000} \textbf{72.32\%}} \\ \hline
\end{tabular}
\end{table*}

\subsection{Results}
\paragraph{Evaluation metrics}
We evaluate our model with two standard metrics: 3D IoU and 2D IoU. The 3D IoU measures the intersection over union between the predicted 3D layout and the ground truth. The 2D IoU measures the intersection over the union between the predicted 2D layout and the ground truth.  

\begin{table}
\centering
\caption{The results on the PanoContext. Our method achieves the best result on PanoContext dataset.}
\label{tab:2}
\begin{tabular}{|c|c|c|}
\hline
Dataset                       & Model                   & 3D IoU (\%)                            \\ \hline 
                              & AtlantaNet\cite{pintore2020atlantanet}             & 78.76                                 \\ \cline{2-3}  
                              & DuLaNet \cite{yang2019dula}                & 77.42                                 \\ \cline{2-3} 
                              & LayoutNet \cite{zou2018layoutnet}               & 74.48                                 \\ \cline{2-3}  
                              & HorizonNet \cite{sun2019horizonnet}             & 82.17                                 \\ \cline{2-3}  
                              & HoHoNet \cite{sun2021hohonet}                & 82.82                                 \\ \cline{2-3}  
                              & LED2  \cite{wang2021led2}                  & {82.75}                               \\ \cline{2-3}  
\multirow{-7}{*}{PanoContext} & Ours                    & {\color[HTML]{FE0000} \textbf{83.88}} \\ \hline
\end{tabular}
\end{table}

\paragraph{Quantitative results}
We exam our model on two datasets and compare it with popular 360 room layout estimation methods, including HorizonNet \cite{sun2019horizonnet}, HoHoNet \cite{sun2021hohonet}, AtlantaNet \cite{pintore2020atlantanet}, and LED2Net \cite{wang2021led2}. Table \ref{tab:1} and Table \ref{tab:2} separately present the reconstruction results on two datasets. Overall, our method achieves the best results on 
Mp3D and PanoContext datasets. Compared with the HorizonNet and HoHoNet methods that also use one-dimensional layout representation, PanoViT achieves an improvement of $2$\% and $1$\% on the Mp3D and PanoContext datasets. This proves the effectiveness of the visual transformer on the room layout estimation task.
Compared with PanoContext, the Mp3D dataset contains more complex room layouts. PanoViT has achieved the best results on the Mp3D dataset (especially in the 10+ corner room layout, PanoViT is 2\% higher than LED2).

\begin{figure*}[t]
\centering
    \includegraphics[width=0.9\textwidth]{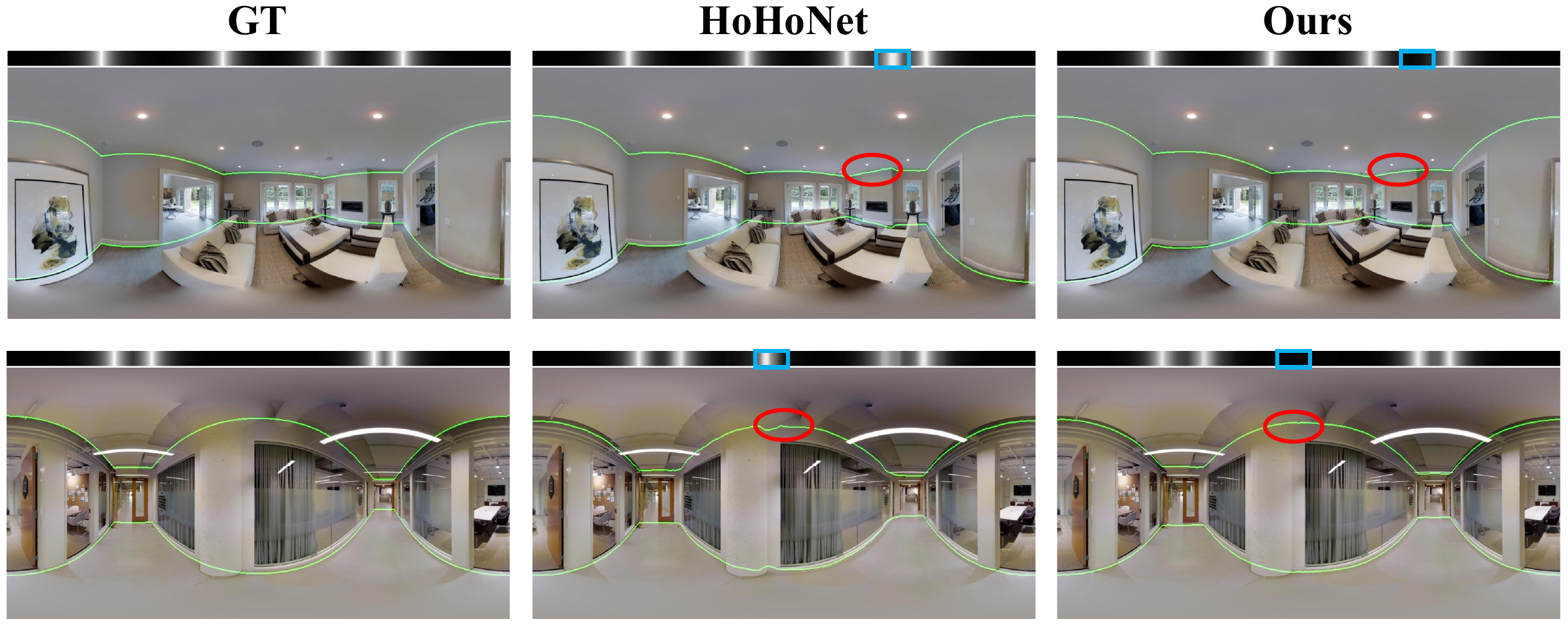}
    \caption{The visualization of the reconstruction result on the Matterport3D dataset.}
    \label{fig:results}
\end{figure*}

\paragraph{Quantitative results}
To better compare the performance of different methods, we also visualize some reconstruction results in Fig. \ref{fig:results}. The results illustrate that PanoViT achieves a better result on the local details.

\begin{table}
\caption{The results of the PanoViT with different inputs. PanoViT w/4-scale uses feature maps of all four blocks from ResNet34, while  PanoViT w/1-scale only uses feature maps of the final blocks and PanoViT w/0-scale ignore the feature maps from the ResNet. }
\centering
\label{tab:3}
\begin{tabular}{|c|c|c|c|}
\hline
\multirow{2}{*}{Dataset}     & \multirow{2}{*}{Model} & \multicolumn{2}{c|}{Metric} \\ \cline{3-4} 
                             &                            & 2D IoU        & 3D IoU        \\ \hline
\multirow{3}{*}{PanoContext} & PanoViT w/4-scale               & \textbf{86.84\%}      & \textbf{83.88\%}     \\ \cline{2-4} 
                              & PanoViT w/1-scale                 & 79.50\%      & 74.94\%      \\ \cline{2-4} 
                             & PanoViT w/0-scale                  & 69.14\%      & 65.32\%      \\ \hline
                            
\end{tabular}
\end{table}

\subsection{Ablation Study}
\paragraph{Multi-scale features}
Besides the patches sampled on the original panoramic image, the PanoViT also receives the patches sampled from the multi-scale feature maps. To exam the effect of the multi-scale feature maps, we test our model with patches sampled from the original image, the patches sampled from both the original image and the feature maps of the last blocks in ResNet34, and the patches sampled from both the original image and feature maps of all four blocks.  All the experiments are conducted on the PanoContext dataset. Table \ref{tab:3} shows the results. We can see that using the multi-scale feature map can significantly improve the estimation accuracy. 
\begin{figure*}[t]
\centering
    \includegraphics[width=0.9\textwidth]{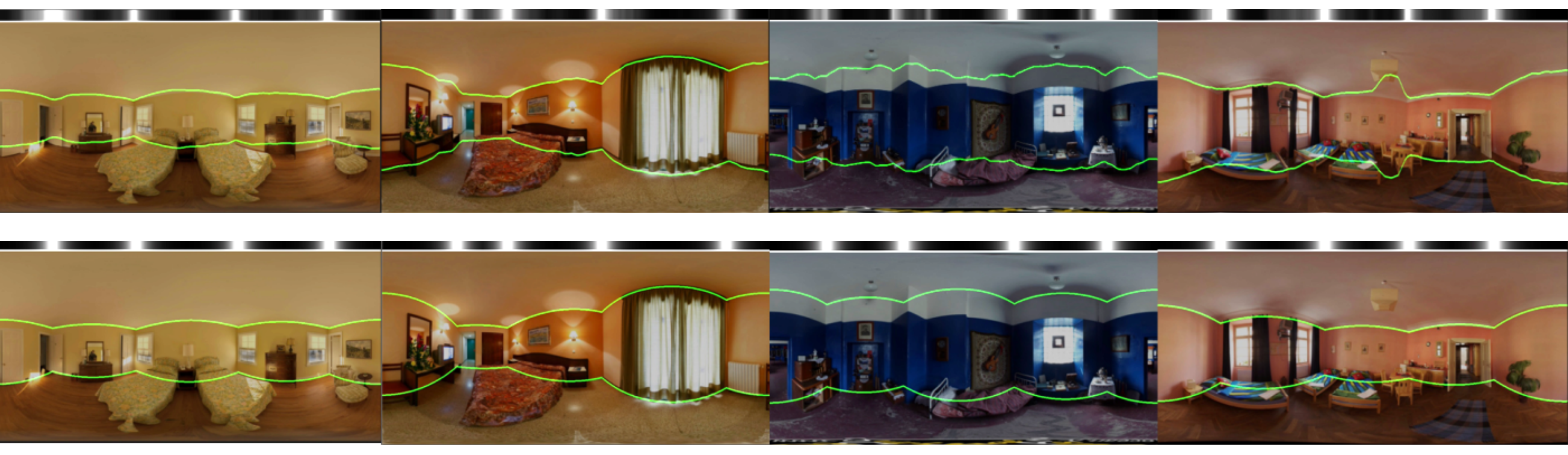}
    \caption{The visualization of reconstruction result of PanoViT with and without CNN multi-scale features. The top row presents the reconstruction result of PanoViT without CNN Multi-scale features. The bottom row presents the reconstruction result with CNN Multi-scale features. The PanoViT with CNN Multi-scale features achieves a better reconstruction result.}
    \label{fig:multi-scale}
\end{figure*}

\paragraph{3D loss, recurrent PE and edge enhancement}

Other ablation experiments are presented in Table \ref{tab:4}. We report the result of our model equipped with different modules, including the 3D loss, the recurrent position embedding, and the edge enhancement module. All the experiments are conducted on the Mp3D dataset. Overall, Table \ref{tab:4} presents that all the modules are helpful for the estimation of the room layout.

\begin{table*}
\caption{The ablation experiments of the PanoViT. The block with $\checkmark$ means that the PanoViT contains the corresponding module.}
\centering
\label{tab:4}
\begin{tabular}{|c|c|c|c|c|ccccc|}
\hline
\multirow{2}{*}{ID} & \multirow{2}{*}{Recurrent PE} & \multirow{2}{*}{Frequency} & \multirow{2}{*}{3D Loss} & \multirow{2}{*}{metric} & \multicolumn{5}{c|}{of   corners (\%)}                                                                                      \\ \cline{6-10} 
                    &                               &                                               &                          &                         & \multicolumn{1}{c|}{overall} & \multicolumn{1}{c|}{4}     & \multicolumn{1}{c|}{6}     & \multicolumn{1}{c|}{8}     & 10+   \\ \hline
1                   &                               &                                               &                          & \multirow{6}{*}{3D IoU} & \multicolumn{1}{c|}{81.49}   & \multicolumn{1}{c|}{84.34} & \multicolumn{1}{c|}{84.09} & \multicolumn{1}{c|}{75.23} & 69.88 \\ \cline{1-4} \cline{6-10} 
2                   &                               &                                               & \checkmark                        &                         & \multicolumn{1}{c|}{81.53}   & \multicolumn{1}{c|}{84.01} & \multicolumn{1}{c|}{83.94} & \multicolumn{1}{c|}{76.14} & 71.01 \\ \cline{1-4} \cline{6-10} 
3                   & \checkmark                             &                                               & \checkmark                        &                         & \multicolumn{1}{c|}{81.70}    & \multicolumn{1}{c|}{\textbf{84.70}}  & \multicolumn{1}{c|}{84.62} & \multicolumn{1}{c|}{76.03} & 69.86 \\ \cline{1-4} \cline{6-10} 
4                   &                               & \checkmark                                             & \checkmark                        &                         & \multicolumn{1}{c|}{81.77}   & \multicolumn{1}{c|}{84.21} & \multicolumn{1}{c|}{84.11} & \multicolumn{1}{c|}{76.16} & 71.93 \\ \cline{1-4} \cline{6-10} 
5                   & \checkmark                             & \checkmark                                             & \checkmark                        &                         & \multicolumn{1}{c|}{\textbf{82.04}}   & \multicolumn{1}{c|}{84.40}  & \multicolumn{1}{c|}{\textbf{84.67}} & \multicolumn{1}{c|}{\textbf{76.34}} & \textbf{72.32} \\ \hline
\end{tabular}
\end{table*}

Experiments 1 and 2 test the effectiveness of the 3D loss. The model in experiment 1 is equipped with 3D loss while the model in experiment 2 is equipped with $L_{1}$ loss. The results in Table \ref{tab:4} show that the model with 3D loss is 0.11\% better than the model with $L_{1}$ loss. 

Experiments 2 and 3, and experiments 4 and 5 show the effectiveness of the recurrent position embedding. The model in experiment 2 is equipped with the learnable position embedding same as \cite{dosovitskiy2020image}, while the model in experiment 3 is equipped with recurrent position embedding. We observe that the model with recurrent position embedding achieves a better 3D IoU result. Experiments 4 and 5 further present the results about the recurrent position embedding when the model is equipped with the edge enhancement module. The model with recurrent position embedding achieves a better result.

Experiments 4 and 5 show the effectiveness of the recurrent position embedding. We test the model with Fourier edge enhancement and the model without Fourier edge enhancement. The results in Table \ref{tab:4} show that the model with edge enhancement shows better performance in all types of room layouts.

\section {Conclusions}
In this paper, we have proposed PanoViT to estimate the room layout from a single panoramic image. PanoViT successfully applied the transformer pre-trained on perspective images to panoramic image processing. The patch sampling method and the recurrent position embedding in PanoViT are proved to be effective for panoramic image processing. The proposed frequency-domain edge enhancement further improves the prediction accuracy. The experimental results on two datasets
validate that the proposed PanoViT can achieve better results.

\bibliographystyle{splncs04}
\bibliography{egbib}

\begin{thebibliography}{10}
\providecommand{\url}[1]{\texttt{#1}}
\providecommand{\urlprefix}{URL }
\providecommand{\doi}[1]{https://doi.org/#1}

\bibitem{brown2020language}
Brown, T.B., Mann, B., Ryder, N., Subbiah, M., Kaplan, J., Dhariwal, P.,
  Neelakantan, A., Shyam, P., Sastry, G., Askell, A., et~al.: Language models
  are few-shot learners. arXiv preprint arXiv:2005.14165  (2020)

\bibitem{carion2020end}
Carion, N., Massa, F., Synnaeve, G., Usunier, N., Kirillov, A., Zagoruyko, S.:
  End-to-end object detection with transformers. In: European Conference on
  Computer Vision. pp. 213--229. Springer (2020)

\bibitem{devlin2018bert}
Devlin, J., Chang, M.W., Lee, K., Toutanova, K.: Bert: Pre-training of deep
  bidirectional transformers for language understanding. arXiv preprint
  arXiv:1810.04805  (2018)

\bibitem{dosovitskiy2020image}
Dosovitskiy, A., Beyer, L., Kolesnikov, A., Weissenborn, D., Zhai, X.,
  Unterthiner, T., Dehghani, M., Minderer, M., Heigold, G., Gelly, S., et~al.:
  An image is worth 16x16 words: Transformers for image recognition at scale.
  In: International Conference on Learning Representations (2020)

\bibitem{esser2021taming}
Esser, P., Rombach, R., Ommer, B.: Taming transformers for high-resolution
  image synthesis. In: Proceedings of the IEEE/CVF Conference on Computer
  Vision and Pattern Recognition. pp. 12873--12883 (2021)

\bibitem{hou2020strip}
Hou, Q., Zhang, L., Cheng, M.M., Feng, J.: Strip pooling: Rethinking spatial
  pooling for scene parsing. In: Proceedings of the IEEE/CVF Conference on
  Computer Vision and Pattern Recognition. pp. 4003--4012 (2020)

\bibitem{kingma2014adam}
Kingma, D.P., Ba, J.: Adam: A method for stochastic optimization. arXiv
  preprint arXiv:1412.6980  (2014)

\bibitem{liu2021swin}
Liu, Z., Lin, Y., Cao, Y., Hu, H., Wei, Y., Zhang, Z., Lin, S., Guo, B.: Swin
  transformer: Hierarchical vision transformer using shifted windows. arXiv
  preprint arXiv:2103.14030  (2021)

\bibitem{pintore2020atlantanet}
Pintore, G., Agus, M., Gobbetti, E.: Atlantanet: Inferring the 3d indoor layout
  from a single 360 image beyond the manhattan world assumption. In: European
  Conference on Computer Vision. pp. 432--448. Springer (2020)

\bibitem{pintore2020state}
Pintore, G., Mura, C., Ganovelli, F., Fuentes-Perez, L., Pajarola, R.,
  Gobbetti, E.: State-of-the-art in automatic 3d reconstruction of structured
  indoor environments. In: Computer Graphics Forum. vol.~39, pp. 667--699.
  Wiley Online Library (2020)

\bibitem{radford2019language}
Radford, A., Wu, J., Child, R., Luan, D., Amodei, D., Sutskever, I., et~al.:
  Language models are unsupervised multitask learners. OpenAI blog
  \textbf{1}(8), ~9 (2019)

\bibitem{sun2019horizonnet}
Sun, C., Hsiao, C.W., Sun, M., Chen, H.T.: Horizonnet: Learning room layout
  with 1d representation and pano stretch data augmentation. In: Proceedings of
  the IEEE/CVF Conference on Computer Vision and Pattern Recognition. pp.
  1047--1056 (2019)

\bibitem{sun2021hohonet}
Sun, C., Sun, M., Chen, H.T.: Hohonet: 360 indoor holistic understanding with
  latent horizontal features. In: Proceedings of the IEEE/CVF Conference on
  Computer Vision and Pattern Recognition. pp. 2573--2582 (2021)

\bibitem{vaswani2017attention}
Vaswani, A., Shazeer, N., Parmar, N., Uszkoreit, J., Jones, L., Gomez, A.N.,
  Kaiser, {\L}., Polosukhin, I.: Attention is all you need. In: Advances in
  neural information processing systems. pp. 5998--6008 (2017)

\bibitem{wang2021led2}
Wang, F.E., Yeh, Y.H., Sun, M., Chiu, W.C., Tsai, Y.H.: Led2-net: Monocular
  360deg layout estimation via differentiable depth rendering. In: Proceedings
  of the IEEE/CVF Conference on Computer Vision and Pattern Recognition. pp.
  12956--12965 (2021)

\bibitem{xu2017pano2cad}
Xu, J., Stenger, B., Kerola, T., Tung, T.: Pano2cad: Room layout from a single
  panorama image. In: 2017 IEEE winter conference on applications of computer
  vision (WACV). pp. 354--362. IEEE (2017)

\bibitem{yang2016efficient}
Yang, H., Zhang, H.: Efficient 3d room shape recovery from a single panorama.
  In: Proceedings of the IEEE conference on computer vision and pattern
  recognition. pp. 5422--5430 (2016)

\bibitem{yang2019dula}
Yang, S.T., Wang, F.E., Peng, C.H., Wonka, P., Sun, M., Chu, H.K.: Dula-net: A
  dual-projection network for estimating room layouts from a single rgb
  panorama. In: Proceedings of the IEEE/CVF Conference on Computer Vision and
  Pattern Recognition. pp. 3363--3372 (2019)

\bibitem{yang2018automatic}
Yang, Y., Jin, S., Liu, R., Kang, S.B., Yu, J.: Automatic 3d indoor scene
  modeling from single panorama. In: Proceedings of the IEEE Conference on
  Computer Vision and Pattern Recognition. pp. 3926--3934 (2018)

\bibitem{zhang2014panocontext}
Zhang, Y., Song, S., Tan, P., Xiao, J.: Panocontext: A whole-room 3d context
  model for panoramic scene understanding. In: European conference on computer
  vision. pp. 668--686. Springer (2014)

\bibitem{SETR}
Zheng, S., Lu, J., Zhao, H., Zhu, X., Luo, Z., Wang, Y., Fu, Y., Feng, J.,
  Xiang, T., Torr, P.H., Zhang, L.: Rethinking semantic segmentation from a
  sequence-to-sequence perspective with transformers. In: CVPR (2021)

\bibitem{zhu2020deformable}
Zhu, X., Su, W., Lu, L., Li, B., Wang, X., Dai, J.: Deformable detr: Deformable
  transformers for end-to-end object detection. In: International Conference on
  Learning Representations (2020)

\bibitem{zou2018layoutnet}
Zou, C., Colburn, A., Shan, Q., Hoiem, D.: Layoutnet: Reconstructing the 3d
  room layout from a single rgb image. In: Proceedings of the IEEE Conference
  on Computer Vision and Pattern Recognition. pp. 2051--2059 (2018)

\bibitem{zou20193d}
Zou, C., Su, J.W., Peng, C.H., Colburn, A., Shan, Q., Wonka, P., Chu, H.K.,
  Hoiem, D.: 3d manhattan room layout reconstruction from a single 360 image.
  arXiv preprint arXiv:1910.04099  (2019)

\end{thebibliography}
\end{document}